\documentclass[letterpaper, 10 pt, conference]{ieeeconf}  

\IEEEoverridecommandlockouts                              

\usepackage{amsmath} 
\usepackage{amssymb}  
\usepackage{caption}
\usepackage{subcaption}
\usepackage{graphicx}
\usepackage{algorithm}
\usepackage{algpseudocode}
\usepackage[utf8]{inputenc}
\usepackage{hyperref}
\providecommand{\keyword}[1]
{
  \small	
  \textbf{\textit{Keywords---}} #1
}

\title{\LARGE \bf
Discrete time model predictive control for humanoid walking with step adjustment }
\author{Vishnu Joshi$^{*,1}$, Suraj Kumar$^{*,1}$, Nithin V$^{1}$ and Shishir Kolathaya$^{1,2}$
\thanks{$^*$These authors contributed equally to this work}
\thanks{$^{1}$Robert Bosch Centre for Cyber Physical Systems, Indian Institute of Science,Bengaluru, India.\{vishnujoshi, surajkumar13, nithinv, shishirk\}@iisc.ac.in}
\thanks{$^{2}$Computer Science and Automation, Indian Institute of Science, Bengaluru, India.\{shishirk\}@iisc.ac.in}
\thanks{We would like to acknowledge Dr S E Talole, Dr Kiran Akella and team of R\&D EE for their suggestions on the robot model. This work was funded by DIA-RCOE, IISc, Bengaluru.}
}

\begin{document}

\maketitle
\thispagestyle{empty}
\pagestyle{empty}

\begin{abstract}

This paper presents a Discrete-Time Model Predictive Controller (MPC) for humanoid walking with online footstep adjustment. The proposed controller utilizes a hierarchical control approach. The high-level controller uses a low-dimensional Linear Inverted Pendulum Model (LIPM) to determine desired foot placement and Center of Mass (CoM) motion, to prevent falls while maintaining the desired velocity. A Task Space Controller (TSC) then tracks the desired motion obtained from the high-level controller, exploiting the whole-body dynamics of the humanoid. Our approach differs from existing MPC methods for walking pattern generation by not relying on a predefined foot-plan or a reference center of pressure (CoP) trajectory. The overall approach is tested in simulation on a torque-controlled Humanoid Robot. Results show that proposed control approach generates stable walking and prevents fall against push disturbances. 

\end{abstract}
\keyword{\small Model Predictive Control (MPC),Task Space Control (TSC), Linear Inverted Pendulum Mode(LIPM)}

\section{INTRODUCTION}
The primary objective of a humanoid robot is to avoid falling while achieving desired locomotion goals, such as walking with desired velocity under external disturbances. Despite having many degrees of freedom, a humanoid robot's balance is primarily determined by the dynamics of the Center of Mass (CoM) relative to the Center of Pressure (CoP). To prevent falls, various recovery strategies can be employed, including ankle torques for CoP modulation, upper body distortions by swinging the arms and torso and recovery steps. Among these balance strategies, recovery step is the most effective means for control. In the event of unforeseen disturbances acting on the robot, it is essential to adjust the stepping location in real-time to prevent the robot from falling. This paper presents a hierarchical control approach that employs a reduced robot model based on the Linear Inverted Pendulum Model (LIPM) within model predictive controllers, enabling online step adaptation in humanoid robots. \\
\begin{figure}[h]
  \centering
  \includegraphics[width=0.3\textwidth]{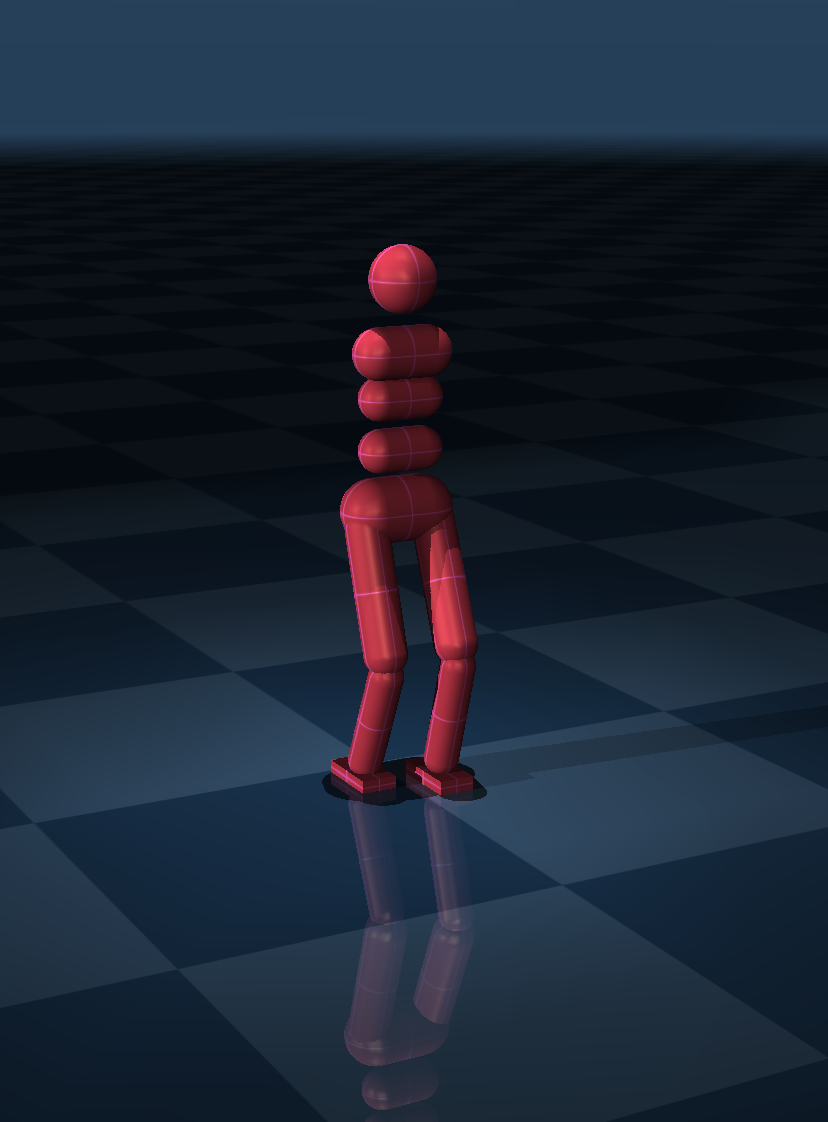}
  \caption{Humanoid robot used for simulation}
  \label{fig:biped}
\end{figure}
Kajita et al. \cite{c13} introduced Linear Inverted Pendulum Model (LIPM) for walking pattern generation where the robot is treated as point mass and constraint to move in plane parallel to ground. Weiber et al. \cite{c1,c2} introduced a continuous time Linear Model Predictive Control for generating CoM trajectories while tracking reference CoP trajectory arising from a pre-defined foot plan and zero moment point constraints. To account for step length adjustment, quadratic cost in deviation of a reference foot plan is added in optimization objective. Task space controller based on whole body dynamics have been proposed in \cite{c3,c4,c5,c7}. When multiple tasks need to be tracked, prioritized optimization is solved to realize desired motion. In \cite{c4,c5,c8} multiple tasks are combined in weighted approach wherein high priority tasks are given higher weights and resultant optimization is solved by weighted Quadratic Programming (QP). In \cite{c6,c7} prioritized approach is considered wherein low priority task is projected in null space of high priority task to ensure that higher priority tasks are not effected. \\
The main contribution of this work is the development of Discrete time Model Predictive controller that uses LIPM model to plan footsteps online. Our approach differs from existing MPC methods for walking pattern generation by not relying on predefined footholds or a reference center of pressure (CoP) trajectory. In conventional MPC approaches, foot placement is generated with the goal of maintaining the desired speed and tracking a reference foot plan. These approaches also generate a CoP trajectory, utilizing ankle torques for balance. However, since ankle torques have limited influence on balance compared to foot placement \cite{c14}\cite{c15}, we do not consider ankle torques for balance within the MPC layer. Instead, CoP modulation is included in TSC as part of contact constraints. \\
The remainder of the paper is organized as follows. Section \ref{sec:prw}
provides necessary background. Section \ref{sec:control} details components of MPC based control framework based on LIPM framework and task space controller formulation based on weighted QP approach. Section \ref{sec:results} shows the results from simulation studies and Section \ref{sec:conc} presents concluding remarks and future works. \\

\section{BACKGROUND}  \label{sec:prw}
\subsection{Humanoid Model}
Humanoid robots are categorized as hybrid dynamical systems, featuring continuous time dynamics followed by discrete ground impact event. Mathematically, hybrid system $\Sigma$ can be defined as \cite{c15}  \\ 
\begin{equation}
    \label{hybrid_dyn}
    \Sigma : \left \{
        \begin{aligned}
        \dot{x} = f(x,u)  \quad \quad  \forall \phi(x) \neq 0, \\
        x^+ = \Delta(x^-) \quad \quad   \forall \phi(x) = 0  \\
        \end{aligned}
    \right \}
\end{equation}
where the continuous time dynamics is given by \(\dot{x} = f(x,u)\) with x and u representing state and control respectively;  \(\phi(x) = 0\) represents the switching or impact condition;  $\Delta$ represents the impact dynamics or impact map. \\

The Humanoid robot used for simulation is a 18 DoF robot with 12 actuated joints (6 DoFs in each leg and 6DoFs in floating base) as show in figure \ref{fig:biped} that weights 69 kg. Detailed specifications and physical properties are listed in table \ref{tab:robot_specs}. For a bipedal robot with $n_v$ degrees of freedom, its continuous time dynamics can be represented by standard multi-body equation with floating base given by,
\begin{equation}
\label{robot_dyn}
\begin{aligned}
    H(q)\ddot{q} + C(q,\dot{q}) \dot{q} + G(q) &= S^T_{a}\tau + J_{s}(q)^T F_{s} \\
\end{aligned}
\end{equation}

\begin{table}[]
    \centering
    \begin{tabular}{|c|c|c|}
    \hline
        \textbf{Robot link} & \textbf{Mass (kg)} & \textbf{Link length (m)} \\
    \hline
        Head & 3.1 & 0.18\\
    \hline
        Torso & 5.9 & 0.14\\
    \hline
        Upper waist & 2.3 & 0.12\\
    \hline
        Lower waist & 2.3 & 0.12\\
    \hline
        Pelvis & 24 & 0.18\\
    \hline
        Thigh & 10 & 0.34\\
    \hline
        Shin & 4 & 0.3\\
    \hline
        Foot & 3 & 0.2\\
    \hline
    \end{tabular}
    \caption{Robot weight}
    \label{tab:robot_specs}
\end{table}

where $q \in R^{n_q}$ is the generalized joint position vector, $\dot{q} \in R^{n_v}$ is the generalized joint velocity vector;$H \in \mathbb{R}^{n_v \times n_v}$, $C \in \mathbb{R}^{n_v}$,$G \in \mathbb{R}^{n_v}$ are mass matrix, coriolis-centripetal term and gravity respectively. Here $F_s \in \mathbb{R}^{6N_{s}}$ is the stacked ground wrench corresponding to $N_{s}$ contact points and $J_s \in \mathbb{R}^{6N_{s} \times n}$ is combined contact Jacobian. $\tau$ is the generalized force vector along joint axes and $S_a$ represents the actuation selection matrix.\\
The humanoid model is constructed in MuJoCo \cite{c11}, Multi body dynamics simulation software and will be used for dynamics and control simulation.

\subsection{Linear Inverted Pendulum Model}
In this work, Linear Inverted Pendulum Model \cite{c13} that captures the evolution of linear momentum of the robot is used as template reduced-order model for model predictive control formulation. In LIPM, angular momentum around Center of Mass (CoM) is assumed to be zero and CoM motion is constrained in plane parallel to ground. This results in linear model for walking which can be solved analytically.\\
Consider the robot motion in 3D plane with X-Z plane representing the sagittal plane and Y-Z plane representing the frontal plane of motion. In LIPM model, dynamics in sagittal plane and frontal plane is decoupled. Let $S = [S_x^T,S_y^T]^T \in R^4$ represent the state of the system where $S_x=[x,\dot{x}]^T$ represent state in sagittal plane and $S_y=[y,\dot{y}]^T$ represents state in frontal plane. Here where $x,y$ and $\dot{x},\dot{y}$ represent CoM position and velocity in sagittal and frontal plane respectively. Let $p = [p_{x},p_{y}] \in R^2$ denote the current support foot position in X-Y plane. Then LIPM model is given as 
\begin{equation}
    \begin{aligned}
        \ddot{x} = \frac{g}{z_0}(x-p_{x}) \\
        \ddot{y} = \frac{g}{z_0}(y-p_{y})
    \end{aligned}
\end{equation}
where, $z_0$ is the height of pendulum, $g$ is the gravitational constant, $x$ and $y$ are the positions of CoM, $p_x$ and $p_y$ are the positions of CoP
Given known timing t, future CoM state at $t_{TD}$ can then be expressed as:
\begin{equation}
\begin{aligned}
    \label{lipm_eqn}
    S_x(t_{TD})  &= A(t_{TD})S_{x}(t) + B(t_{TD})p_x \\
    A(t) &= 0.5\begin{bmatrix}
(e^{\omega t} + e^{-\omega t}) & \frac{(e^{\omega t} - e^{-\omega t})}{\omega} \\
\omega (e^{\omega t} - e^{-\omega t}) & (e^{\omega t} + e^{-\omega t})
\end{bmatrix}   \\
    B(t) &= \begin{bmatrix}
1 - 0.5 (e^{\omega t} + e^{-\omega t}) \\
-0.5 \omega (e^{\omega t} - e^{-\omega t})
\end{bmatrix}
\end{aligned}
\end{equation}
where $A(t)$ is the state transition matrix, $B(t)$ is the control map, $\omega = \sqrt{g/z_0}$ represents the natural frequency of the model and $t_{TD}$ denotes the touchdown timing of swing leg. Similar expression hold also for frontal plane as dynamics is decoupled and identical. This forms the basis of discrete time Model Predictive Control as state equation can be propagated from one physical step to another step. 

\section{Humanoid Control}  \label{sec:control}
\begin{figure}[h]
  \centering
  \includegraphics[width=0.5\textwidth]{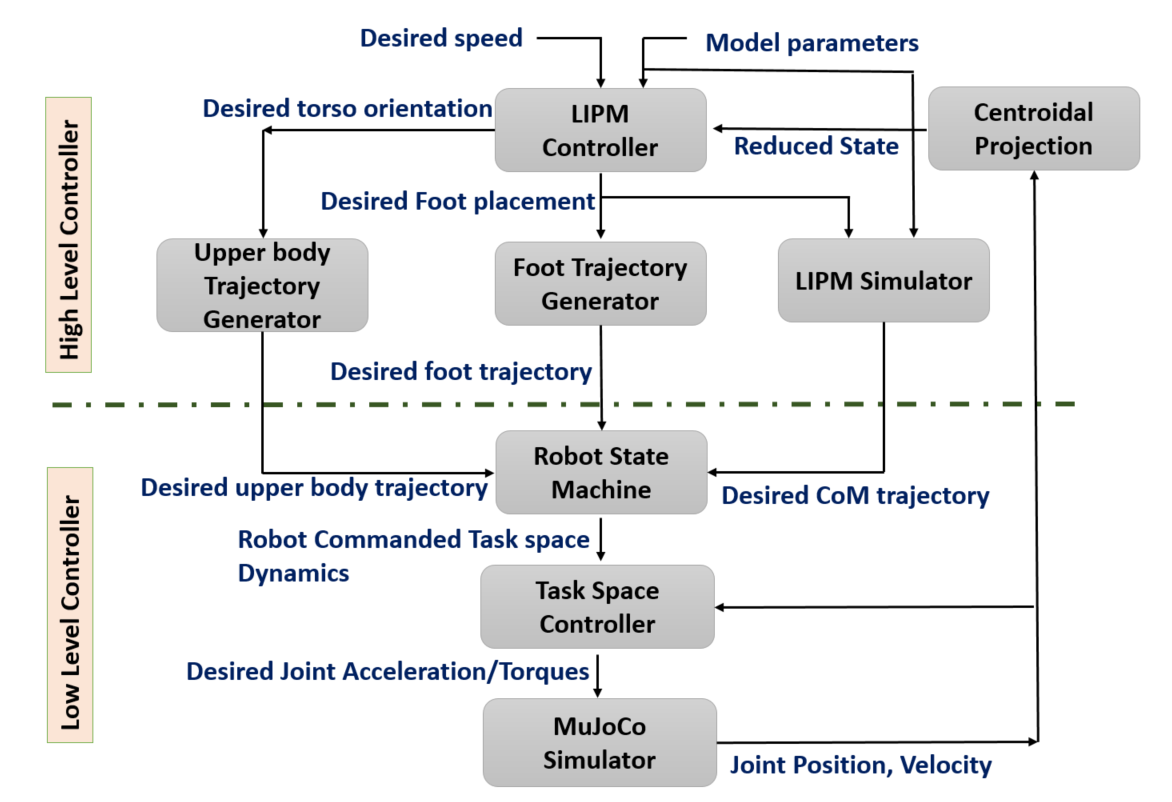}
  \caption{Humanoid Control Block Diagram}
  \label{fig:lipm_block_diagram}
\end{figure}

The controller comprises of two layers as shown in figure \ref{fig:lipm_block_diagram}. High level controller uses LIPM to plan CoM trajectories and footsteps using linear discrete time Model Predictive Control. The low level controller uses optimization based task space control for tracking task space trajectories generated by high level controller.  

\subsection{Discrete Time Model Predictive Control} 
Our first contribution is the development of a foot placement controller for maintaining balance while tracking desired velocity without use of pre-defined foot-holds. We assume instantaneous double support phase and consider step timing to be constant to keep MPC formulation linear which can be solved reliably fast on embedded computer via Quadratic Program (QP). It should be noted that finite double support phase can also be accommodated in same formulation as long as the phase duration is kept constant or planned ahead with zero acceleration. \\
We now present the convex optimization based formulation for generating foot placement to maintain balance and track desired CoM velocity. We note that equation \ref{lipm_eqn} can be propagated for N physical steps owing to analytical dynamics as follows:
\begin{equation}
\begin{aligned}
    \label{lipm_n_eqn}
    S_1 & = \bar{A}(t_{TD})S_0 + \bar{B}(t_{TD})p_0 \\
    S_2 & = \bar{A}(T)S_1 + \bar{B}(T)p_1 \\
    S_3 & = \bar{A}(T)S_2 + \bar{B}(T)p_2 \\
    S_4 & = \bar{A}(T)S_3 + \bar{B}(T)p_3 \\
    & \vdots \\
    S_N & = \bar{A}(T)S_{N-1} + \bar{B}(T)p_{N-1}
\end{aligned}
\end{equation}
where $\bar{A}$ and $\bar{B}$ is obtained by from LIPM system matrix $A, B$ as follows
\begin{equation}
    \bar{A}(t) =  \begin{bmatrix} A(t) & 0 \\
    0 & A(t) \\
\end{bmatrix}, 
 \bar{B}(t) =  \begin{bmatrix} B(t) & 0 \\
    0 & B(t) \\
\end{bmatrix}
\end{equation}
Current support foot position, $p_0$ is fixed and cannot be changed but future foot placements \([p_{1},p_{2},...,p_{N}]\) can be optimized to achieve balance while tracking desired CoM motion. The cost function $J$ for optimization can be specified as quadratic cost in state and control as
\begin{equation}
\begin{aligned}    
    J &= \sum_{k=1}^{N-1} ((S_{k+1} - S^*)^T Q (S_{k+1}-S^*) \\
      &+ (p_{k+1}-p_k)^T R (p_{k+1}-p_k)) 
\end{aligned}
\end{equation}
For the purposes of walking, user defined center of mass velocity is the desired center of mass trajectory and cost function penalizes the deviation of desired velocity. In this approach, we choose not to specify desired center of mass position which evolves as per the dynamics of the robot. To do so, entries in Q matrix associated with position are made 0. We also don’t assume availability of desired foot plan and relative foot placement is penalized in quadratic cost. This
form also ensures asymptotic stability in LIPM dynamics as minimizing any derivative of the motion of the CoM of the robot results in stable online walking motion \cite{c1}. From another perspective, minimizing the difference in foot placement is equivalent to minimizing the average CoM velocity. Closer foot steps result in less fluctuation of CoM velocity during each single support phase, which contributes to generating a more stable walking motion. The foot placement controller can then be formulated as QP:
\begin{equation}
\begin{aligned}    
    \underset{p_1,p_2,...,p_N}{\text{minimize}} 
    & \sum_{k=1}^{N-1} ((S_{k+1} - S^*)^T Q (S_{k+1}-S^*) \\
    & + (p_{k+1}-p_k)^T R (p_{k+1}-p_k))  
\end{aligned}
\end{equation}
subject to dynamics constraints given by eq \ref{lipm_n_eqn} and bounding box constraint:
\begin{equation}
     l_i \leq p_{i+1} - p_{i} \leq  u_i \forall i=2,....,N
\end{equation}
where $l_i$ and $u_i$ represents the foot placement constraints

\subsection{Task Space Control}
The goal of task space controller is to track the desired motions generated by high level controller subject to dynamics constraints, unilateral ground reaction constraints and actuation constraints. The desired motions, referred to as task variables henceforth include key quantities like CoM motion, centroidal angular momentum, swing leg motions etc. In our design, desired task variables are CoM desired trajectory, swing leg trajectory and torso trajectory. \\
Each of these task variables are expressed as linear function of joint space variables as,
\begin{equation}
r_t = \mathbf{A}_t \dot{q}
\end{equation}
where \(\mathbf{A}_t\) is the generalized task Jacobian. The task space dynamics is then given as,
\begin{equation}
\mathbf{A}_t\Ddot{q} + \mathbf{\dot{A}}_t\dot{q} = \dot{r_t}
\label{ts_dyn}
\end{equation}
\subsubsection{Commanded task space dynamics}
The commanded task space dynamics is imposed as linear control law of the form,
\begin{equation}
    \dot{r}_{t,c} = \Ddot{p}_d + K_d(\dot{p}_d - \dot{p}) + K_p(p_d - p)
\end{equation}
Here, \(\Ddot{p}_d,\dot{p}_d, p_d\) are the desired acceleration, desired velocity and position and \(\dot{p}, p\) are the current velocity and position associated with the given task and $K_p, K_d$ are the tunable proportional and derivative gain associated with the task. For linear momentum task the following is used,
\begin{equation}
    \dot{r}_{t,c} = \dot{l}_{G,d} + K_d(\dot{p}_{G,d} - \dot{p}_G) + K_p(p_{G,d} - p_G)
\end{equation}
where $\dot{l}_{G,d},\dot{p}_{G,d}, p_{G,d}$ are desired rate of change of linear momentum, CoM velocity,  CoM position respectively and $\dot{p}_G, p_G$ are the current velocity and position of the CoM respectively. For angular momentum only a damping term is used as follows,
\begin{equation}
    \dot{r}_{t,c} = \dot{k}_{G,d} + K_d(k_{G,d} - k_G)
\end{equation}

where $\dot{k}_{G,d}, k_{G,d}$ are the desired rate of change of angular momentum, angular momentum respectively and $k_G$ is the current angular momentum.

\subsubsection{Control formulation}
The goal of controller is to compute the desired contact forces, joint accelerations \(\Ddot{q}\), and joint torques \(\tau\)  that minimizes the tracking error in the task space dynamics subject to dynamics constraints. The resultant minimization problem is formulated as QP, 
\begin{equation}
\begin{aligned}
    \label{tsc_formulation}
    & \underset{\ddot{q},\tau,F_{s}}{\text{minimize}} 
    & & ||A_{t}(q)\ddot{q} + \dot{A_{t}}(q) \dot{q} - \dot{r}_{t,c}||^2_{W_{t}} + ||\tau||^2_{R_{\tau}} + ||F_{s}||^2_{R_{\lambda}} \\
    & \text{subject to} 
    & & H(q)\ddot{q} + C(q,\dot{q}) + G(q) = S^T_{a}\tau + J_{s}(q)^T F_{s} \\ 
    & & & F_{s} \in \mathcal{P} \\ 
    & & & |\tau| < \tau_{max}
\end{aligned}
\end{equation}

where \(\mathcal{P}\) is the friction cone constraint approximated as pyramid constraint. It is given by,
\begin{equation}
\mathcal{P} = \left\{ (n_x, n_y, n_z, f_x, f_y, f_z) \in \mathbb{R}^6 \ \middle|\ 
\begin{aligned}
& f_x \leq \mu f_z, \\
& f_y \leq \mu f_z, \\
& d_x^- \leq \frac{n_y}{f_z} \leq d_x^+, \\
& d_y^- \leq \frac{n_x}{f_z} \leq d_y^+
\end{aligned}
\right\}
\end{equation}

where \(\mu, d_x, d_y\) are the friction coefficient, half the length of the foot along x and y axis. Here \(\tau_{min}, \tau_{max}\) are joint torque constraints. \(W_i, Q, R\) are the weight matrix associated with state and control penalty. \\
The optimization \ref{tsc_formulation} is formulated as a QP and is then implemented using qpSWIFT \cite{c10}.

\section{Results}  \label{sec:results}
\begin{figure}[h]
  \centering
  \includegraphics[width=0.5\textwidth]{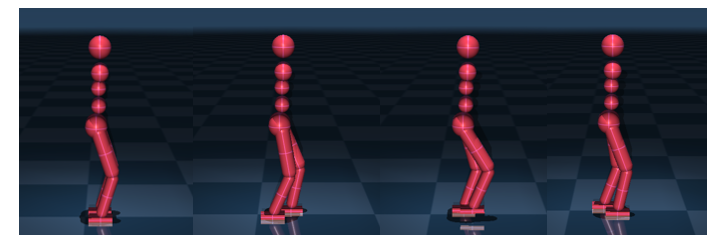}
  \caption{Humanoid Walking Gait}
  \label{fig:lipm_walking}
\end{figure}

\begin{figure}[h!]
  \centering
  \includegraphics[width=0.4\textwidth]{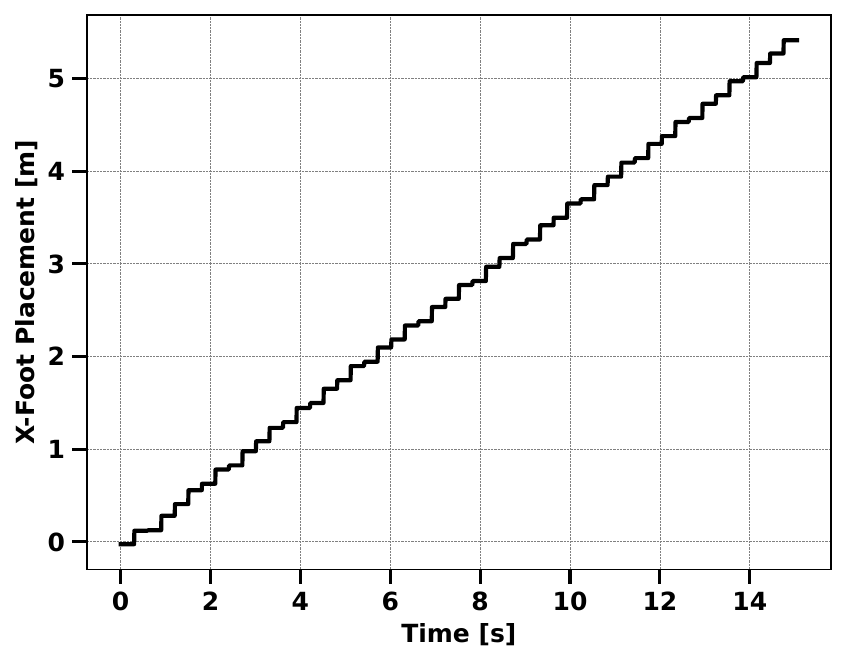}
  \caption{Foot Placement}
  \label{fig:foot_placement}
\end{figure}

\begin{figure}[h!]
  \centering
  \includegraphics[width=0.45\textwidth]{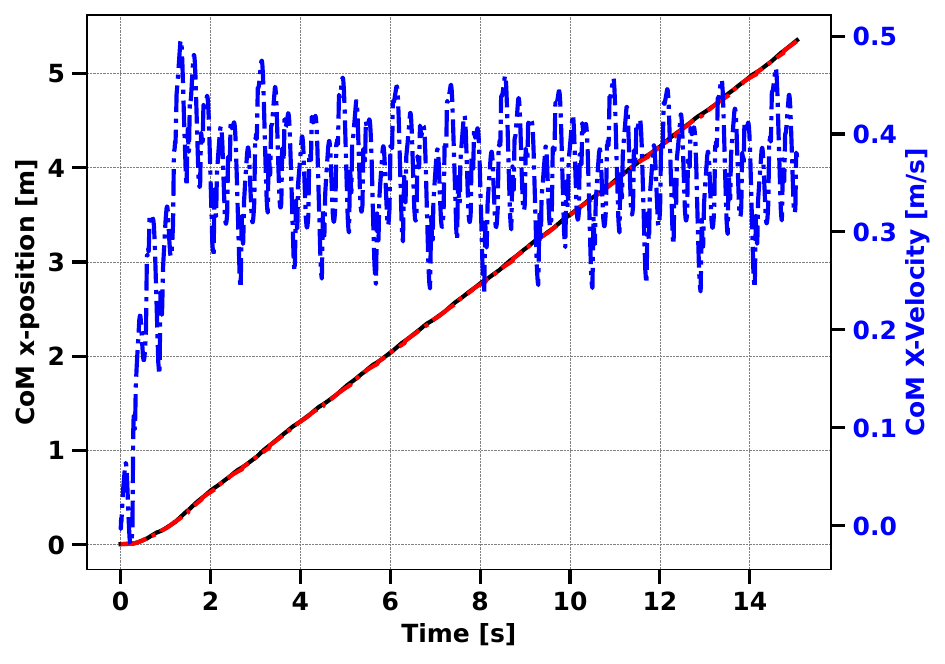}
  \caption{CoM x-position and velocity}
  \label{fig:com_tracking}
\end{figure}

\begin{figure}[h!]
  \centering
  \includegraphics[width=0.4\textwidth]{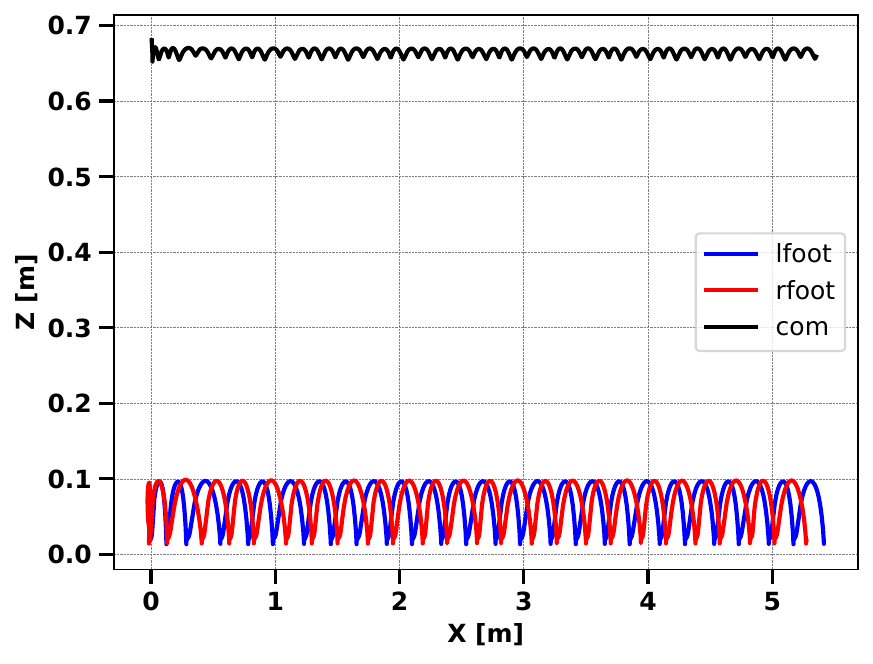}
  \caption{CoM and Foot Trajectory trace obtained from TSC}
  \label{fig:traj_trace}
\end{figure}

Figure \ref{fig:lipm_block_diagram} shows standard implementation block diagram of LIPM locomotion controller. Foot placement controller computes next foot placement based on desired walking speed which is fed to foot trajectory generator module that computes desired foot trajectory from current foot position to target foot position. Foot placement and model parameters are used by LIPM simulator to compute center of mass trajectory. In LIPM, upper body orientation is regulated to vertical as prescribed by the LIPM equation of motion. This is fed to upper body trajectory generator module to generate torso orientation trajectory. Robot state machine then selects appropriate task dynamics for the foot, CoM and upper body to continuously track the RoM-CoM dynamics and realize the desired foot touchdown locations. Joint space controller then selects whole-body joint torques/joint accelerations at real-time to realize these desired tasks using task space controller. Figure \ref{fig:lipm_walking} shows stable walking gait obtained from this implementation with forward commanded velocity of 0.4 m/s. Figure \ref{fig:foot_placement} shows the foot placement obtained from MPC implementation. Figure \ref{fig:com_tracking} shows evolution of CoM trajectory. MPC controller takes approximately 3 physical steps to acquire desired commanded velocity. Foot placement also reaches near steady state taking 0.1m per steps exhibiting near periodic motion. Error in tracking CoM and foot trajectory reflects as disturbance in high level MPC controller which is compensated by adjustment of step length around steady state step length. Currently, MPC is called at the start of single support phase once per step. Tracking error can be further reduced by increasing the MPC call frequency. Figure \ref{fig:traj_trace} shows the trajectory trace of CoM and swing foot obtained from TSC.

\subsection{Walking with payload}
\begin{figure}[h!]
  \centering
  \includegraphics[width=0.4\textwidth]{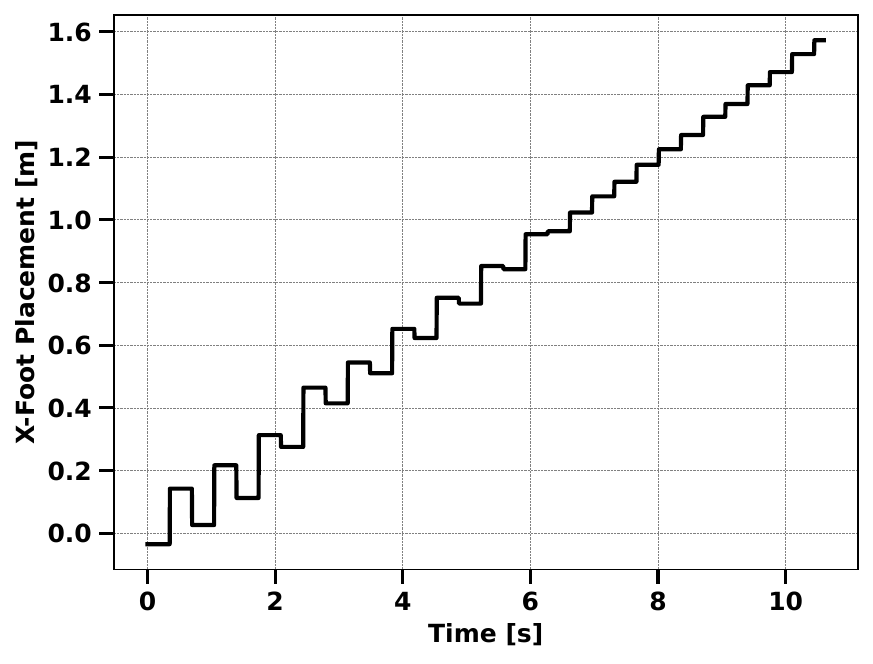}
  \caption{Foot Placement under payload variations}
  \label{fig:foot_placement_payload}
\end{figure}

\begin{figure}[h!]
  \centering
  \includegraphics[width=0.45\textwidth]{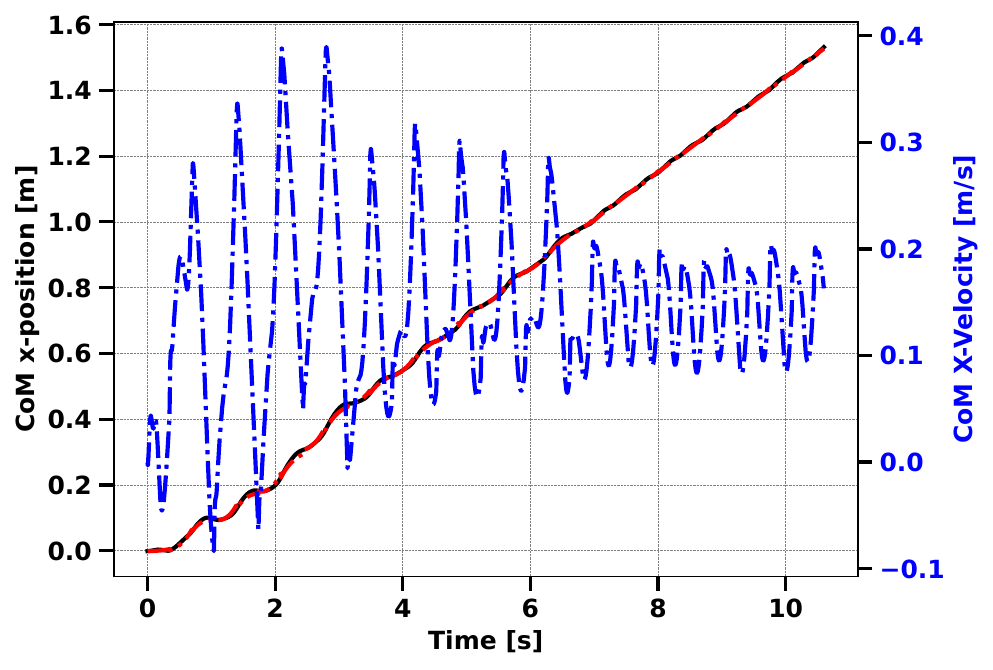}
  \caption{CoM evolution under payload variations}
  \label{fig:com_tracking_payload}
\end{figure}

\begin{figure}[h!]
  \centering
  \includegraphics[width=0.4\textwidth]{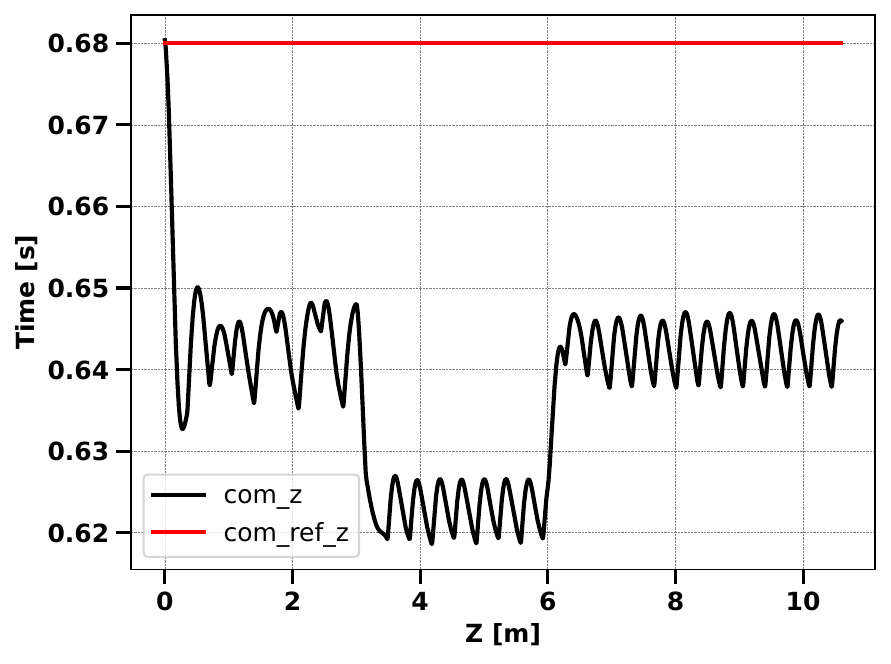}
  \caption{CoM z-tracking under payload variations}
  \label{fig:com_z_tracking_payload}
\end{figure}

Simulation experiments are conducted for different values of payload carrying capabilities. Controller demonstrated stable and robust walking while carrying payload upto 30 kg for 3s starting at 3s in the simulation with forward commanded velocity of 0.2 m/s. Figure \ref{fig:foot_placement_payload} and  shows the MPC generated reactive foot placement to account for payload variations and Figure \ref{fig:com_tracking_payload} shows CoM evolution. Effect of payload variation is observed in the increment of steady state error in z-position tracking in CoM as we only use PD controller in task space dynamics as evident in figure \ref{fig:com_z_tracking_payload}  

\subsection{Walking under push disturbance}
\begin{figure}[h!]
  \centering
  \includegraphics[width=0.4\textwidth]{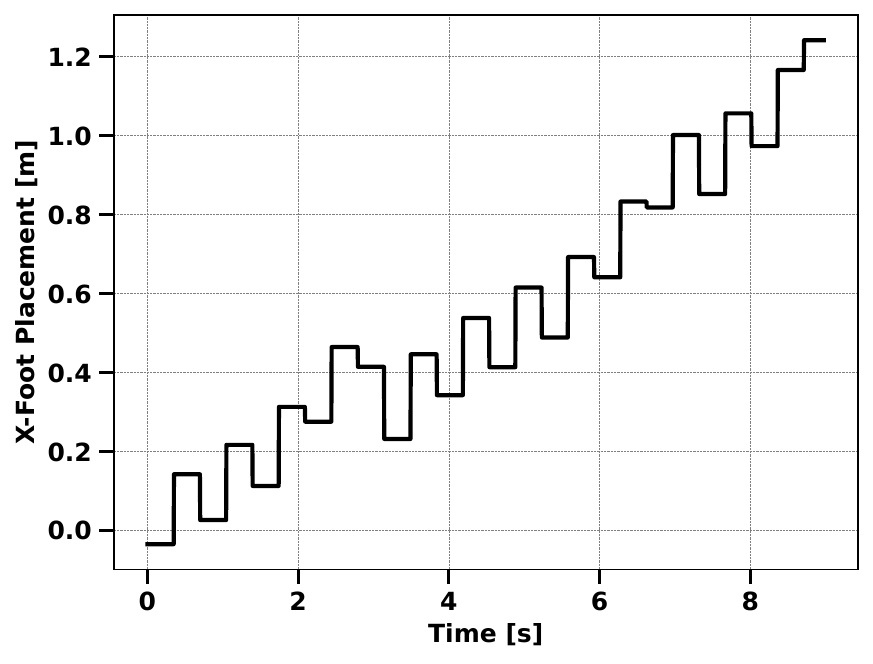}
  \caption{Foot Placement under push disturbance}
  \label{fig:foot_placement_disturbance}
\end{figure}

\begin{figure}[h!]
  \centering
  \includegraphics[width=0.45\textwidth]{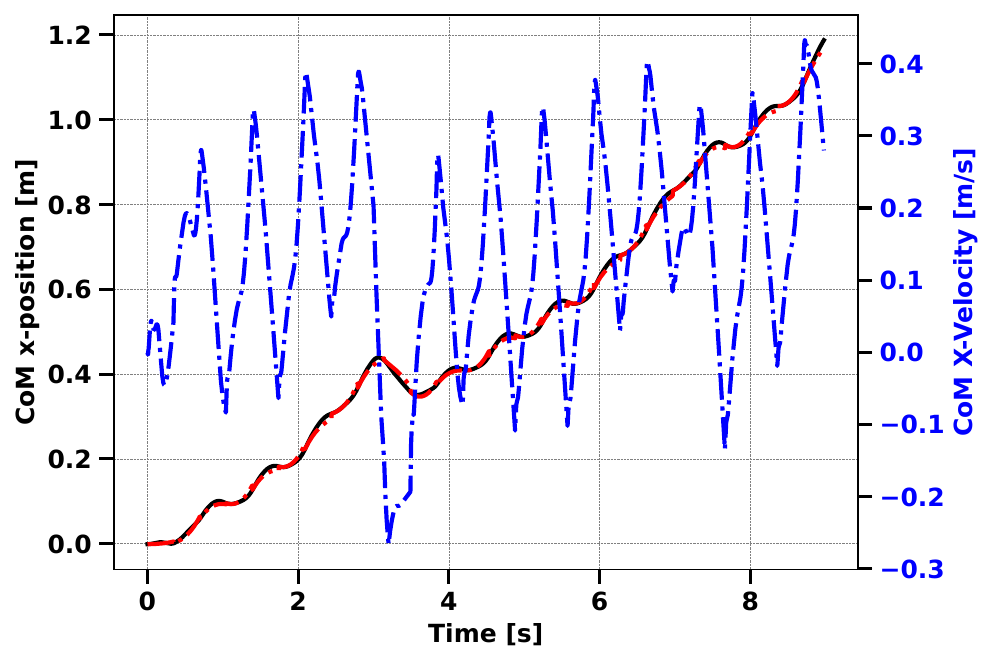}
  \caption{CoM evolution under push disturbance}
  \label{fig:com_tracking_disturbance}
\end{figure}

Simulation experiments are conducted to establish controller robustness against external push disturbances of magnitude 150N is applied to robot's lower waist for 0.2s in the opposite direction of robot velocity vector with a commanded velocity of 0.2m/s. Figure \ref{fig:foot_placement_disturbance} shows the reactive foot placement under push disturbance. Figure \ref{fig:com_tracking_disturbance} shows the CoM evolution under push disturbance.

When push disturbance is applied at 3s, CoM velocity decreases to -0.2 m/s due to deceleration by external disturbance. MPC absorbs the disturbance by taking reactive steps in backward direction and prevents fall. After 3-4 recovery steps, the momentum due to external disturbance is completely absorbed and robot eventually reaches steady state commanded velocity of 0.2 m/s. 

\subsection{4cm stairs proprioceptive walking}
\begin{figure}[h!]
  \centering
  \includegraphics[width=0.4\textwidth]{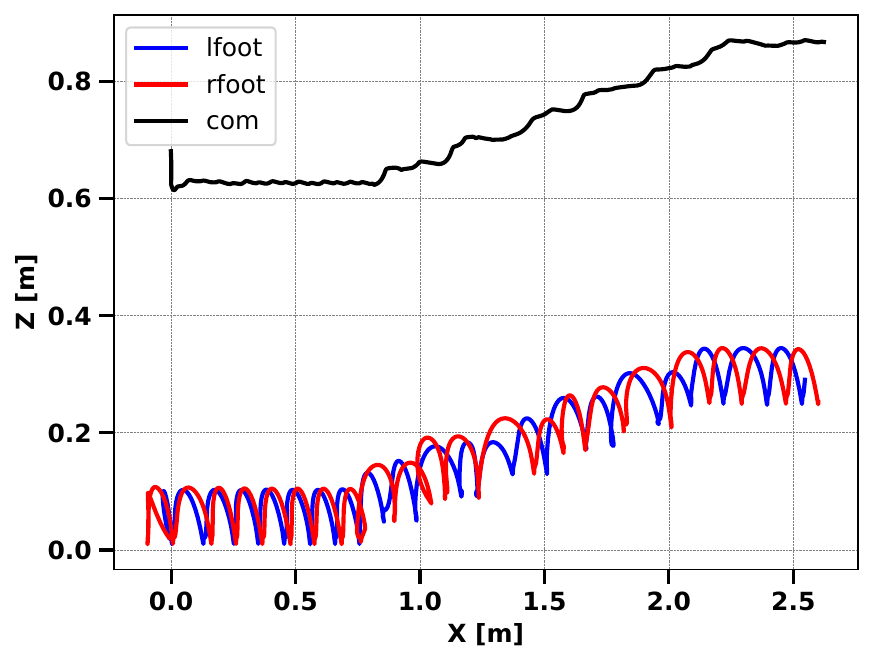}
  \caption{CoM and foot trajectory from 4cm stairs proprioceptive walking}
  \label{fig:4cm_stairs_traj_trace}
\end{figure}

\begin{figure}[h!]
  \centering
  \includegraphics[width=0.45\textwidth]{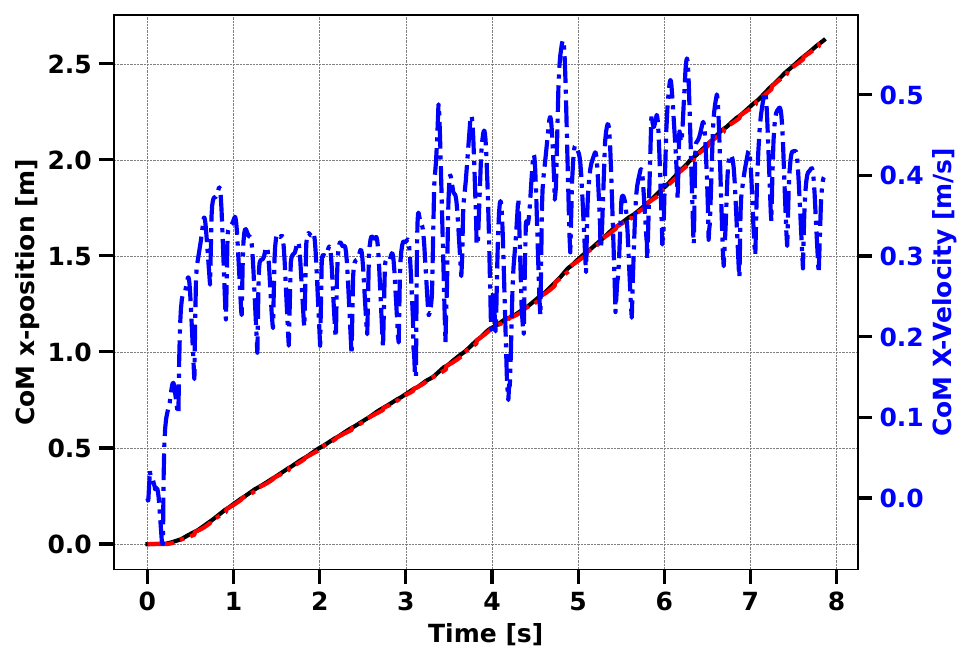}
  \caption{CoM evolution from 4cm stairs proprioceptive walking}
  \label{fig:4cm_stairs_com_tracking}
\end{figure}

Simulation experiments are conducted to establish controller capabilities in proprioceptive walking with 4 cm stairs. Figure \ref{fig:4cm_stairs_traj_trace} shows the height trajectory evolution of swing foot and center of mass as it climbs the stairs. Figure \ref{fig:4cm_stairs_com_tracking} shows the evolution of CoM as it tracks desired forward speed of 0.4 m/s. Error in tracking is due to interference of task dynamics in weight QP based task space controller. Figure \ref{fig:4cm_stairs} shows image of humanoid traversing 4cm stairs.

\subsection{Unknown terrain proprioceptive walking}
\begin{figure}[h!]
  \centering
  \includegraphics[width=0.4\textwidth]{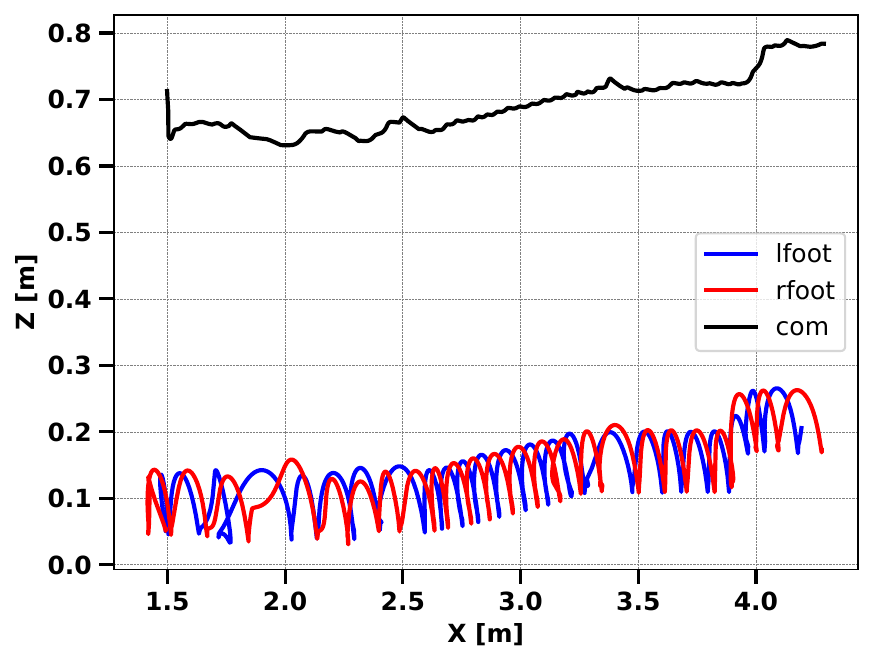}
  \caption{CoM and foot trajectory from unknown terrain proprioceptive walking}
  \label{fig:unknown_terrain_traj_trace}
\end{figure}

\begin{figure}[h!]
  \centering
  \includegraphics[width=0.45\textwidth]{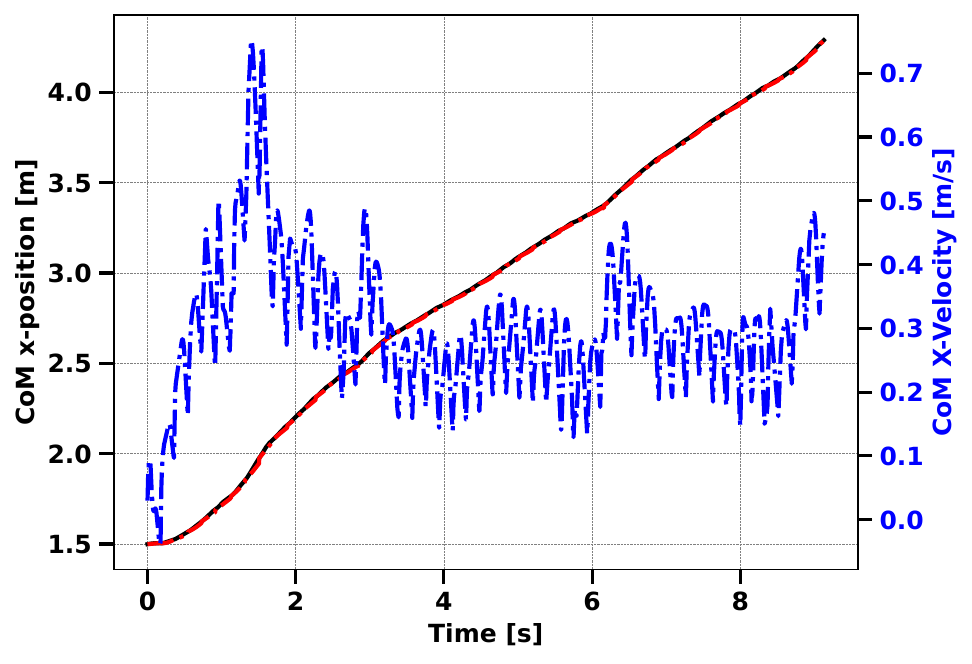}
  \caption{CoM evolution from unknown terrain proprioceptive walking}
  \label{fig:unknown_terrain_com_tracking}
\end{figure}

\begin{figure}[h!]
  \centering
  \includegraphics[width=0.45\textwidth]{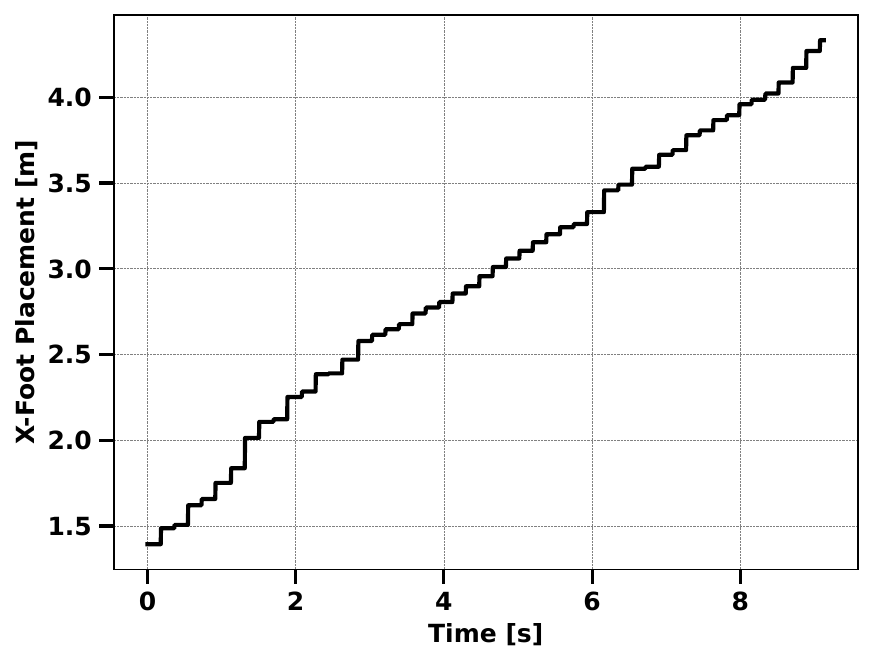}
  \caption{Foot placement on unknown terrain}
  \label{fig:unknown_terrain_fp}
\end{figure}

\begin{figure}[h!]
  \centering
  \includegraphics[width=0.45\textwidth]{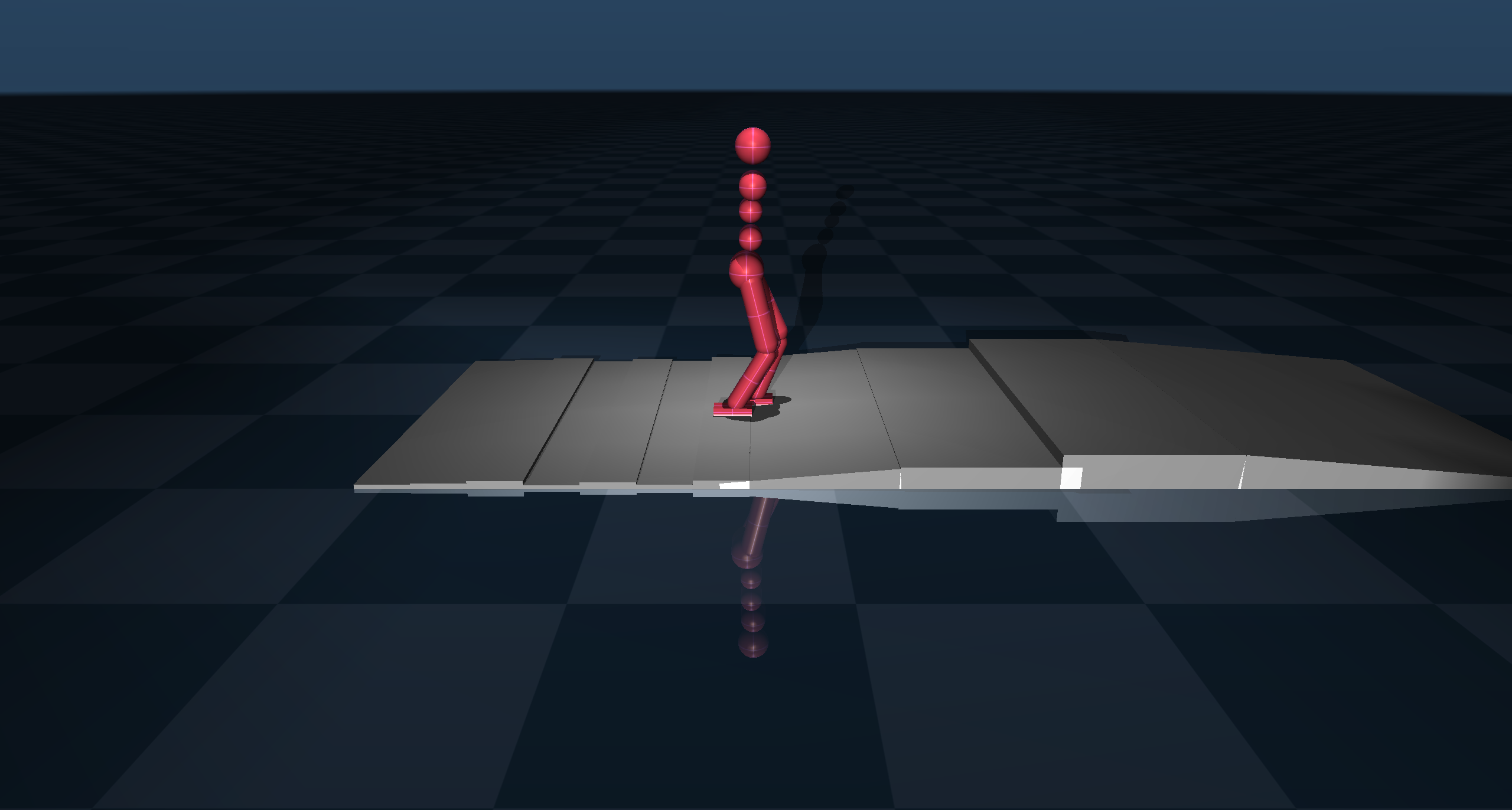}
  \caption{Humanoid walking on unknown terrain}
  \label{fig:unknown_terrain_walking}
\end{figure}

Simulation experiments are conducted to establish controller capabilities in proprioceptive walking with random 4 cm height variations. Figure \ref{fig:unknown_terrain_traj_trace}-\ref{fig:unknown_terrain_com_tracking} shows the sagittal plane trajectory evolution and CoM position and velocity respectively. Figure \ref{fig:unknown_terrain_walking} shows the image of the humanoid walking on unkonwn terrain. Figure \ref{fig:unknown_terrain_fp} shows the reactive foot placement by MPC to absorb the terrain disturbances while tracking desired velocity of 0.4 m/s. 

Video demonstration of stable walking for each case can be accessed \href{https://youtu.be/D4F4fwbIQbg}{here}.

\section{Conclusion and future work}  \label{sec:conc}
The proposed MPC controller has proven to be effective in generating stable walking under push disturbance, payload variations and rough terrains by adjusting step length on the fly. However, we observed that weighted QP formulation in TSC requires extensive tuning and often produced interference in high priority tasks by low priority tasks through weights tuning. This resulted in limited basin of attraction for range of commanded velocities with which robot can walk. We observed that tuning gains and weights for walking with $>$ 1m/s turned out to be notoriously difficult. Infact, interference between task dynamics leads to tracking error which is reflected as disturbance in MPC. This problem can be alleviated by increasing the MPC call frequency. \\
In future, we would first like to reformulate the weighted QP based task space optimization to prioritized task space optimization to robustify the joint level controller and demonstrate multi-directional walking. We will also extend this framework for slope and rough terrain walking. Finally, the assumptions in LIPM model can be further relaxed by augmenting it with learning based controllers which would certainly improve walking gait and range of walking speed.

\end{document}